\documentclass[]{fairmeta}

\usepackage[utf8]{inputenc} 
\usepackage[T1]{fontenc}    
\usepackage{url}            
\usepackage{booktabs}       
\usepackage{caption}
\usepackage{graphicx}
\usepackage{amssymb}
\usepackage{amsmath}
\usepackage{amsthm}
\usepackage{amsfonts}       
\usepackage{nicefrac}       
\usepackage{microtype}      
\usepackage{algorithm}
\usepackage[endLComment=,italicComments=false]{algpseudocodex}
\usepackage{xcolor, colortbl}
\usepackage{tabularx}
\usepackage{multirow}
\usepackage{caption}
\usepackage{subcaption}
\usepackage{wrapfig}
\definecolor{lightcyan}{rgb}{0.88, 1, 1}
\usepackage[title]{appendix}
\usepackage{titlesec}
\usepackage{soul}  

\usepackage{tikz}
\usetikzlibrary{patterns}

\theoremstyle{plain}

\theoremstyle{definition}

\theoremstyle{remark}

\newlength\savewidth

\makeatletter

\newcommand{\customdashline}[3]{%
    \noalign{\vbox{\hrule height 0pt%
        \hbox to\linewidth{\cleaders\hbox to \dimexpr #1 + #2\relax{\hss\rule{#1}{#3}\hss}\hfill}%
    }}%
}
\makeatother

\renewcommand{\paragraph}[1]{\vspace{-.3em}\noindent\textbf{#1}}
\newcolumntype{x}[1]{>{\centering\arraybackslash}p{#1pt}}
\newcolumntype{y}[1]{>{\raggedright\arraybackslash}p{#1pt}}
\newcolumntype{z}[1]{>{\raggedleft\arraybackslash}p{#1pt}}
\newcommand{\app}{\raise.17ex\hbox{$\scriptstyle\sim$}}

\definecolor{deemph}{gray}{0.58}

\definecolor{baselinecolor}{gray}{.9}

\newcommand{\RNum}[1]{\uppercase\expandafter{\romannumeral #1\relax}}

\definecolor{lightcyan}{rgb}{0.88, 1, 1}
\definecolor{asparagus}{rgb}{0.53, 0.66, 0.42}
\definecolor{azure}{rgb}{0.0, 0.5, 1.0}
\definecolor{brightpink}{rgb}{1.0, 0.0, 0.5}
\definecolor{boston}{rgb}{0.8, 0.0, 0.0}
\definecolor{gray}{rgb}{0.75, 0.75, 0.75}
\definecolor{lightgray}{rgb}{0.88, 0.88, 0.88}
\definecolor{darkgray}{rgb}{0.50, 0.50, 0.50}
\definecolor{orange2}{rgb}{0.99, 0.86, 0.70}
\definecolor{pastelgray}{rgb}{0.81, 0.81, 0.77}
\definecolor{orangered}{rgb}{.99, .40, .00}
\definecolor{darkUT}{HTML}{BF5700}
\definecolor{lightUT}{HTML}{FFF1E6}
\definecolor{darkUT_B}{HTML}{005F86}
\definecolor{lightUT_B}{HTML}{F5FDFF}
\definecolor{darkUT_G}{HTML}{333F48}
\definecolor{lightUT_G}{HTML}{F5F4F0}
\definecolor{darkgreen}{HTML}{F5F4F0}
\definecolor{apricot}{HTML}{FFD7B8}

\newcommand{\motivbox}[2][]{%
  \tikz[baseline=(text.base)] \node[
    fill=orangered!30,
    inner sep=1pt,
    #1
  ] (text) {#2};
}

\usepackage[most]{tcolorbox}
\newtcolorbox{remarkbox}[1][]{
  enhanced,
  breakable,
  colback=utlight!10,        
  colframe=utlight,             
  colbacktitle=utmain!80!black,         
  coltitle=white,                 
  fonttitle=\bfseries,            
  title=#1,                       
  boxed title style={
    frame empty,
    left=2pt,
    right=2pt,
    top=2pt,
    bottom=2pt,
  },
  rounded corners,                
  boxrule=1pt,                    
  arc=1mm,                        
  before skip=1em,                
  after skip=1em,                 
}

\usepackage{tikz}

\newcounter{takeawayonly}

\newcommand{\parag}[1]{\vspace{+0.0mm}\noindent\textbf{#1}}
\newcommand{\takeawayonly}[1]{
    \vspace{-0.05cm}
    \refstepcounter{takeawayonly}
    \begin{tcolorbox}[
        colback=darkUT!8,                       
        colframe=darkUT!95,                     
        arc=4pt,                    
        boxsep=5pt,                 
        left=2pt,                  
        right=2pt,                 
        top=4pt,                    
        bottom=4pt,                 
        boxrule=0.8pt,              
        drop shadow=gray!30!white,  
        enhanced jigsaw             
    ]
    \vspace{-0.15cm}
        \parag{\textbf{\textit{Remark\,\thetakeawayonly:}}} #1
    \vspace{-0.15cm}
    \end{tcolorbox}
}

\newcounter{closing}
\renewcommand{\parag}[1]{\vspace{+0.0mm}\noindent\textbf{#1}}
\newcommand{\closing}[1]{
    \vspace{-0.05cm}
    \refstepcounter{closing}
    \begin{tcolorbox}[
        colback=darkUT!8,                       
        colframe=darkUT!95,                     
        arc=4pt,                    
        boxsep=5pt,                 
        left=2pt,                  
        right=2pt,                 
        top=4pt,                    
        bottom=4pt,                 
        boxrule=0.8pt,              
        drop shadow=gray!30!white,  
        enhanced jigsaw             
    ]
    \vspace{-0.15cm}
        \parag{\textbf{\textit{Final Remark:}}} #1
    \vspace{-0.15cm}
    \end{tcolorbox}
}

\definecolor{mlBg}{HTML}{FAFAFA}      
\definecolor{mlFg}{HTML}{37474F}      
\definecolor{mlKeyword}{HTML}{7C4DFF} 
\definecolor{mlConst}{HTML}{D81B60}   
\definecolor{mlString}{HTML}{91B859}  
\definecolor{mlComment}{HTML}{90A4AE} 
\definecolor{mlBuiltin}{HTML}{00BCD4}
\definecolor{mlLib}{HTML}{00897B}    
\definecolor{mlLineno}{HTML}{B0BEC5}

\usepackage[T1]{fontenc}
\usepackage[scaled=0.9]{inconsolata}
\newcommand{\pyfont}{\ttfamily}
\usepackage{listings}
\usepackage{caption}
\lstdefinestyle{py-material-light}{
  language=Python,
  backgroundcolor=\color{mlBg},
  basicstyle=\pyfont\small\color{mlFg},
  showstringspaces=false,
  breaklines=true,
  tabsize=4,
  keepspaces=true,
  columns=fullflexible,
  numbers=none, 
  commentstyle=\itshape\color{mlComment},
  stringstyle=\color{mlString},
  keywordstyle=\bfseries\color{mlKeyword},
  morekeywords=[2]{True,False,None},
  keywordstyle=[2]\bfseries\color{mlConst},
  emph={print,range,len,enumerate,zip,dict,list,set,tuple,int,float,str,bool,open,
        sorted,sum,min,max,any,all,abs,super,isinstance,type,property},
  emphstyle=\color{mlBuiltin},
  emph={[2]{np,pd,plt,torch,tf,jax,sklearn}},
  emphstyle=[2]\color{mlLib}
}

\newcommand{\UsePyMaterialLight}{\lstset{style=py-material-light}}

\definecolor{aoBg}{HTML}{FAFAFA}
\definecolor{aoFg}{HTML}{383A42}
\definecolor{aoComment}{HTML}{A0A1A7}

\usepackage[cjk]{kotex}
\usepackage[font=small]{subcaption}
\usepackage[font=footnotesize]{subcaption}

\usepackage[dvipsnames]{xcolor}
\usepackage{lipsum}
\usepackage{adjustbox}
\usepackage{minted} 

\usepackage{listings}

\title{A Recipe for Stable Offline Multi-agent Reinforcement Learning}

\author[1]{Dongsu Lee}
\author[2]{Daehee Lee}
\author[1\dagger]{Amy Zhang}

\affiliation[1]{University of Texas at Austin}
\affiliation[2]{Sungkyunkwan University}
\contribution[\dagger]{Supervision}

\abstract{
Despite remarkable achievements in single-agent offline reinforcement learning (RL), multi-agent RL (MARL) has struggled to adopt this paradigm, largely persisting with on-policy training and self-play from scratch. One reason for this gap comes from the instability of \textit{non-linear} value decomposition, leading prior works to avoid complex mixing networks in favor of linear value decomposition (\textit{e.g.}, {VDN}) with value regularization used in single-agent setups. In this work, we analyze the source of instability in non-linear value decomposition within the offline MARL setting. Our observations confirm that they induce value-scale amplification and unstable optimization. To alleviate this, we propose a simple technique, scale-invariant value normalization (\texttt{SVN}), that stabilizes actor-critic training without altering the Bellman fixed point. Empirically, we examine the interaction among key components of offline MARL (\textit{e.g.}, value decomposition, value learning, and policy extraction) and derive a practical recipe that unlocks its full potential.
}

\date{\today}

\metadata[Correspondence]{Dongsu Lee at \url{dongsu.lee@utexas.edu}}
\metadata[Blog]{\url{https://dongsuleetech.github.io/blog/lazy-borrowed-recipe/}}

\begin{document}
\maketitle
\setcounter{tocdepth}{1}
\addtocontents{toc}{\protect\setcounter{tocdepth}{-1}}

\vspace{+1em}
\section{Introduction}
\label{sec: intro}
\begin{figure}[t]
    \centering
    \includegraphics[width=0.8\columnwidth]{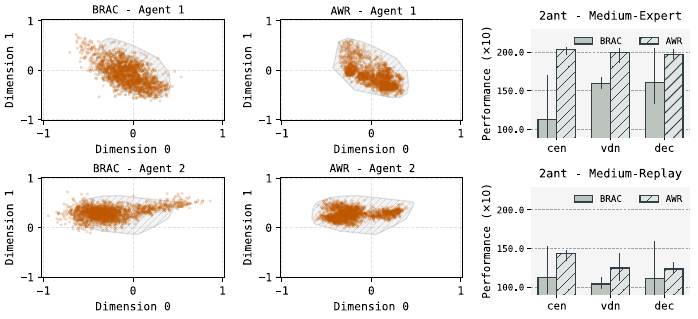}
    \caption{\textbf{Revisiting offline RL insights in MARL}. (\textit{Left}) The convex hull denotes the dataset action support, and dots represent actions sampled from the learned policy. \texttt{BRAC} exhibits mode-seeking behavior that extends beyond the dataset support, while \texttt{AWR} remains mode-covering and strictly in-distribution. (\textit{Right}) Although such mode-seeking would be helpful in single-agent RL, even small out-of-distribution actions induced by \texttt{BRAC} lead to severe performance degradation in MARL, highlighting the sensitivity of joint behavior to individual policy deviations. These results are based on TD learning and hold it regardless of the value decomposition methods (\texttt{cen}tralized, \texttt{vdn}, and \texttt{dec}entralized).}
    \label{fig: obs}
\end{figure}

While offline RL has achieved notable success, its extension to multi-agent settings remains relatively underexplored. More importantly, insights that hold in single-agent settings often fail to transfer to MARL. For example, while DDPG+BC (BRAC)~\citep{fujimoto2021minimalist} can be even preferable to advantage-weighted regression (AWR)~\citep{peng2019advantage} as a single-agent policy extraction~\citep{park2024value}, we observe that in MARL, even minor deviations can precipitate severe performance degradation (\textit{Figure}~\ref{fig: obs}). This instability highlights a key challenge to multi-agent systems: even a minor deviation in individual agent actions can cascade into a complete breakdown of the coordination.

Despite these structural challenges, most existing studies have merely extended single-agent value regularization techniques to multi-agent settings~\citep{wang2023offline, shao2023counterfactual} with linear value decomposition~\citep{li2025dof, lee2025multi} or centralization~\citep{wang2023offline}. Although effective in some cases, centralized critics raise scalability concerns, and linear value decomposition can struggle to capture complex coordination structures. That is, achieving limited short-term success but offering little insight into the deeper challenges of offline MARL. This naturally raises a central question motivating this research: 
\begin{center}
\textit{Where does the bottleneck in offline MARL come from, \\ and how should we design algorithms to explicitly address it?}
\end{center}
In this study, we move beyond simple value regularization with linear value decomposition. Specifically, we first focus on non-linear value decomposition~\citep{rashid2020monotonic} in offline MARL. Then, we empirically investigate the interplay between value decomposition, policy extraction, and value learning to unlock the potential of offline MARL.

Our key discoveries and contributions are as follows:

\motivbox{\textbf{Analyses}} In our pathological observations, the bottleneck of the non-linear method manifests itself as a coupled instability between value learning and policy extraction. Unlike linear decomposition, mixing networks structurally couple per-agent approximation errors through their Jacobian. This coupling breaks the contractivity of the global TD operator and turns value updates expansive rather than contractive. As a result, joint Q-values can grow exponentially even on expert datasets. This value-scale amplification further propagates to policy extraction, where actor gradients become dominated by the absolute magnitude of the value function rather than relative advantages. This dominance leads to poorly calibrated loss and unstable updates. Comprehensively, these effects form a feedback loop that destabilizes learning in non-linear value decomposition.

\motivbox{\textbf{Solutions}} To alleviate this issue, we propose a simple yet effective normalization technique that stabilizes nonlinear value decomposition without altering the Bellman fixed point. Our method renders both critic and actor updates scale-invariant, thereby directly addressing the amplification mechanism induced by the non-linear value decomposition while preserving theoretical correctness. This normalization restores stable and well-scaled optimization dynamics for actor-critic training with nonlinear value decomposition, enabling the non-linear value decomposition to be used reliably in the offline setting for the first time.

\motivbox{\textbf{Experiments}} Our empirical results further offer clear guidance for designing offline MARL algorithms. We find that performance is far more sensitive to value decomposition and policy extraction than value learning methods. In particular, non-linear value decomposition and mode-covering policy extraction yield stable and strong performance. Moreover, we demonstrate that nonlinear value decomposition equipped with our solution is practical across both continuous and discrete control, and remains stable when transitioning from offline to online. These findings highlight value decomposition as both a fundamental bottleneck and a promising lever to advance offline MARL.

\vspace{+1em}
\section{Related work}
\label{sec: related}

Over the past few years, data-driven approaches have fundamentally reshaped control systems, enabling learning-based methods to tackle real-world problems across several domains, including autonomous driving~\citep{liu2023datasets, lee2024ad4rl, lee2024episodic, lee2025episodic} and robotics~\citep{black2024pi_0, o2024open}. A major driver of this progress has been the development of off-policy RL algorithms, which train value functions via temporal difference (TD) learning while leveraging static operational logs~\citep{levine2020offline}.

Compared to single-agent RL, offline MARL faces a more severe out-of-distribution challenge. A small deviation in an individual policy can induce joint behaviors that are absent from the dataset. This can lead to unseen coordination patterns even when each agent's action is individually plausible. To prevent such a failure mode, prior works have largely followed the success of single-agent RL in terms of value regularization and policy extraction, such as conservatism~\citep{pan2022plan, shao2023counterfactual, eldeeb2024conservative}, in-sample maximization~\citep{wang2023offline}, action support constraint~\citep{yang2021believe, jiang2021offline}, distribution matching~\citep{zhu2024madiff, li2025dof, li2025om2p}, convex duality~\citep{matsunaga2023alberdice}, and density weighting~\citep{lee2025multi}. In practice, these approaches rely on behavior-regularized policy gradients(\textit{e.g.,} BRAC~\citep{fujimoto2021minimalist, tarasov2023revisiting}) or AWR~\citep{peng2019advantage} for policy extraction. However, the role of value decomposition remains underexplored. Value decomposition in prior work is limited to simple linear forms (\textit{e.g.}, VDN~\citep{sunehag2017value}) or a fully centralized critic. Although these choices improve stability, they also limit expressivity or introduce scalability challenges as the number of agents grows. More expressive non-linear mixing architectures are often avoided due to long-observed instability~\citep{shen2022resq, liu2025offline}.

Rather than proposing another algorithmic variant, this work aims to diagnose why non-linear value decomposition becomes unstable. We trace this instability to the structural coupling between per-agent value learning and policy extraction that amplifies joint out-of-distribution errors. Based on this analysis, we introduce a simple normalization method that preserves the Bellman fixed point and enables practical non-linear value decomposition. Finally, we study how standard policy extraction methods interact with value learning objectives and decomposition strategies, and distill effective design principles for offline MARL.

\section{Background}
\label{sec: background}
\textbf{Problem formulation.} 
We formulate the MARL problem as a decentralized partially observable Markov decision process (Dec-POMDP)~\citep{oliehoek2016concise} $\mathcal M = \langle \mathcal A, \mathcal S, \{\mathcal U_a\}, \{\mathcal O_a\}, \mathcal P, \Omega, R, \gamma \rangle$. Each agent identified by $a \in \mathcal A \equiv \{1, \cdots, A\}$ takes an action $u^a \in \mathcal U_a$ at a given local observation $o^a \in \mathcal O_a$. The joint action is denoted by $\mathbf{u} = (u^a, \ldots, u^a) \in \mathcal U = \times_a \mathcal U_a$, and the environment evolves according to the transition probability $\mathcal P(s' \mid s, \mathbf{u})$, where $s, s' \in \mathcal S$ represent the global states. After the transition, each agent receives an individual observation $o_i$ sampled from the joint observation function $\Omega(\mathbf{o} \mid s', \mathbf{u})$, where $\mathbf{o} = (o^1, \ldots, o^A) \in \mathcal O = \times_a \mathcal O_a$. Each agent receives an individual reward $r_a(s, u^a, u^{-a}) \in \mathbb{R}$, where $u_{-i}$ denotes the actions of all other agents except agent $i$; that is, the team reward $R(s,\mathbf{u})= \frac{1}{A}\sum_{a=1}^{A} r_a(s, u^a, u^{-a})$ is defined as the aggregation of individual rewards. Each agent acts according to a local policy $\pi_a(u^a \mid \tau^a)$, conditioned on its individual trajectory $\tau^a = (o_0^a, u_0^a, \ldots, o_t^a)$.

Our objective is to identify and address the challenges of offline MARL. We aim to learn a set of policies $\pi = \{\pi^1, \ldots, \pi^A\}$ that maximizes the expected discounted return $\mathbb{E}_{\tau \sim p^{\pi}(\tau)}\left[\sum_{t=0}^{H} \gamma^t R(s_t, \mathbf{u}_t)\right]$ for all $\tau$ from the dataset $\mathcal D$ collected from behavioral policy $\mu$, where $\gamma \in [0,1)$ is the discount factor. The objective of offline MARL is therefore to infer a performant set of decentralized policies $\pi$ while maintaining coordinated behavior across agents under partial observability. This formulation naturally supports centralized training with decentralized execution (CTDE)~\citep{zhang2021multi, gronauer2022multi}. 

\textbf{Value decomposition via the mixing network.} 
To enable CTDE, we adopt value decomposition through a mixing network~\citep{rashid2020monotonic, son2019qtran}. Each agent maintains an individual utility function $Q^a(\tau^a, u^a)$. The global action–value function is represented as follows:
\begin{equation}
    Q_{\mathrm{tot}}(s, \mathbf{u}) = f_{\mathrm{mix}}\big(Q^1, \ldots, Q^A; s\big),
\label{eq: mixer}
\end{equation}
where $f_{\mathrm{mix}}$ is a differentiable function parameterized by a hypernetwork with the global state $s$. To ensure consistency between local and global optima, we impose the monotonicity constraint $\partial Q_{\mathrm{tot}} / \partial Q^a \ge 0$, which guarantees that the joint greedy action can be obtained by independent maximization over each $Q^a$~\citep{rashid2020monotonic}. The network parameters are optimized by minimizing the TD loss,
\begin{equation}
\mathbb{E}_{\substack{(s, \mathbf{u}, r, s’) \sim \mathcal{D} \\ \mathbf{u’} \sim \boldsymbol\pi}} \left[ \left(Q_{\text{tot}}(s, \mathbf u) - r - \gamma \bar{Q}_{\text{tot}}(s’, \mathbf {u’})^2  \right)\right],   
\label{eq: tdloss}
\end{equation}
allowing the mixing network to learn non-linear joint interactions among agents while preserving decentralized policies.

\textbf{Multi-agent actor-critic framework.} 
Building on the decomposed Q-function, we formulate an off-policy actor-critic scheme that factorizes the joint policy through the same mixing structure used for value decomposition~\citep{wu2019behavior}. The global critic $Q_{\mathrm{tot}}(s, \mathbf{u})$ aggregates per-agent utilities by Equation~\eqref{eq: mixer} while maintaining differentiability with respect to each agent’s action. Each actor $\pi^a$ is trained via behavioral regularized actor-critic as follows: 
\begin{equation}
    \mathbb{E}_{s \sim \mathcal{D}}\left[-Q^a(o^a, u^a)+\alpha \underbrace{h\big(\pi^a(u^a|o^a), \mu^a(u^a|o^a)\big)}_{{\color{darkUT_G}\text{behavioral regularization}}} \right],
    \label{eq: brac}
\end{equation} 
where $\alpha$ is a weight coefficient, $h(\cdot, \cdot)$ represents the function that captures the divergence between $\pi^a$ and $\mu^a$. Its gradients are back-propagated from the global critic to each local actor via $f_{\text{mix}}$. This structure implicitly decomposes the policy optimization objective, aligning decentralized policies with the joint value landscape estimated by the centralized critic~\citep{sunehag2017value, wang2020qplex, peng2021facmac}. The resulting formulation enables consistent off-policy updates and coordinated learning without requiring access to global information during execution.

\textbf{{\color{darkUT}Notational warning:}} All policy and critic are parameterized: $\pi_\theta =\{\pi_{\theta_1},\ldots, \pi_{\theta_A}\}$ and $Q^{\text{tot}}_\phi=\{Q_{\phi_1},\ldots,Q_{\phi_A},f_{\phi_{\text{mix}}}\}$. For simplicity, we omit explicit dependence on $\theta$ and $\phi$ in the main text, \textit{i.e.}, writing $\pi$, $Q_{\mathrm{tot}}$, and $f_{\text{mix}}$, whenever no ambiguity arises.


\section{Divergent dynamics of mixer optimization}
\label{sec: divergent}
Off-policy value learning in the actor-critic algorithm is tricky because each Bellman update regresses toward a target that may itself be biased by approximation error or extrapolation. This difficulty is further compounded by nonlinear coupling among per-agent critics through the mixer $f_{\text{mix}}$, which aggregates local utilities into a single joint value. In this section, we first illustrate the necessity of non-linear value decomposition $f_{\text{mix}}$ through a didactic example (Sec.~\ref{subsec: motiv}). Then, we investigate the origin of instability within the mixer (Sec.~\ref{subsec: prob1} and~\ref{subsec: prob2}) and motivate a scale-invariant regularization to restore stability (Sec.~\ref{subsec: fullremedy}). All analyses of the following sub-sections are conducted on a 2ant task with the Expert dataset, except Sec.~\ref{subsec: motiv}.

\begin{figure}[h]
    \centering
    \includegraphics[width=0.6\columnwidth]{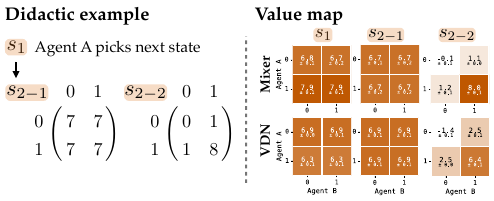}
    \caption{\textbf{Two step matrix game with offline dataset.} (\textit{Left}) The schematic of the didactic example. In $s_1$, Agent A selects between a safe state $s_{2-1}$ with a fixed suboptimal reward of $7$ and a risky state $s_{2-2}$ with an optimal reward of $8$. (\textit{Right}) The learned joint Q value matrices for each state. The top and bottom rows display the linear method (\texttt{VDN}) and Mixer. Each cell reports the mean Q value and two standard deviations across five random seeds.}
    \label{fig: didactic}
    \vspace{-0.1cm}
\end{figure}

\subsection{Didactic example: The necessity of non-linear value decomposition for MARL}
\label{subsec: motiv}

To demonstrate why non-linear value decomposition is necessary, we show a simple didactic example where the linear method (\texttt{VDN}), \textit{i.e.}, $Q_{\text{tot}}(s,\mathbf{u}) = \sum_{a=1}^A Q_a(\tau^a,u^a)$, fails. We adopt the two-step game where agents must choose a safe sub-optimal state and a risky optimal state~\citep{xu2023dual, rashid2020weighted}. For training an offline policy, we augment the dataset to ensure an equal distribution of all possible patterns. \textit{Figure}~\ref{fig: didactic} shows that the \texttt{VDN} fails to represent the non-monotonic payoff structure of the risky state, resulting in an underestimated value and a convergence to the suboptimal safe policy. On the other hand, Mixer can identify the global optimum, and such an observation underscores that non-linear expressivity is a structural necessity for solving complex coordination tasks.

\subsection{Problem I: Coupled value updates break the contractivity of TD learning}
\label{subsec: prob1}

\begin{figure}[h]
    \centering
    \includegraphics[width=0.6\columnwidth]{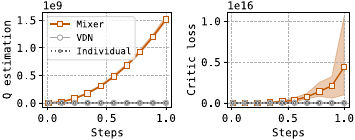}
    \caption{\textbf{Divergent dynamics of mixer-based critics}. Comparison among the monotonic Mixer, VDN, and individual critics under expert offline data. The mixer induces co-amplification of the Q value (\textit{Left}) and critical loss (\textit{Right}), indicating a structural instability of the TD operator.}
    \label{fig: prob1}
\end{figure}

\textbf{Pathological observation.} Primarily, we begin with a controlled comparison between three variants of value decomposition: (\textit{i}) individual critics, (\textit{ii}) the linear value decomposition (\texttt{VDN}), and (\textit{iii}) the nonlinear value decomposition (\texttt{Mixer}). Using an offline dataset of expert demonstrations, we eliminate other components (\textit{e.g.}, exploration and replay stochasticity) so that learning dynamics reflect only the intrinsic behavior of the value updates. As shown in \textit{Figure}~\ref{fig: prob1}, both the individual critics and \texttt{VDN} remain numerically stable throughout training: their total value estimates and TD losses converge smoothly to bounded levels. In contrast, the mixer-based critic exhibits structural divergence: the magnitude of the Q value grows exponentially, and the critic loss co-amplifies by several orders of magnitude. This divergence persists across random seeds, indicating that it arises from the structural coupling of value updates.

\textbf{Analytical formulations.} Define $Q_{\mathrm{tot}}(s,\mathbf{u})$ as in Equation~\eqref{eq: mixer} with Jacobian $J_s = \partial f_s / \partial Q$, where $f_s \triangleq f_\text{mix}(s, \cdot)$ denotes the state-conditioned mixing function. Linearizing $f_{\text{mix}}$ around the current estimate yields the following:
$$Q_{\mathrm{tot}}
\leftarrow Q_{\mathrm{tot}} - 2\alpha_Q (I-\gamma J_s)
\big(Q_{\mathrm{tot}} - \bar Q_{\mathrm{tot}}\big).$$
If $\gamma |J_s|_{\mathrm{op}} > 1$, then $\rho(I-2\alpha_Q(I-\gamma J_s))>1$, and the map becomes expansive, where $|\cdot|_{\mathrm{op}}$ is operator norm $\max_{||x||=1}||J_sx||$ and $\rho(M)$ is spectral radius $\max_i |\lambda_i(M)|$.\footnote{$x$ and $M$ is a placeholder variable and matrix. Then, $\lambda_i(\cdot)$ means $i$-th eigenvalue of a specific matrix.} That is, both $|Q_{\mathrm{tot}}|$ and $\mathcal{L}_{\mathrm{TD}}$ (Equation~\eqref{eq: tdloss}) grow geometrically. With an actor, the closed-loop gain $g_{\mathrm{loop}} = \gamma|J_s|_{\mathrm{op}}
\Big|\frac{\partial \pi}{\partial Q_{\mathrm{tot}}}\Big|$ introduces positive feedback when $g_{\mathrm{loop}}>1$, joint divergence occurs.

\takeawayonly{The mixer’s Jacobian couples per-agent errors into a non-contractive TD operator. Actor updates amplify this coupling, converting TD updates from damping to amplifying.}

\subsection{Problem II: Loss miscalibration under value-scale amplification}
\label{subsec: prob2}

In addition to the instability discussed in Sec~\ref{subsec: prob1}, value-scale amplification in $Q_{\mathrm{tot}}$ leads to further learning issues. A drift in the critic's scale could miscalibrate the policy gradient, causing its magnitude to depend on the absolute value scale instead of action quality. This section analyzes how this value-scale amplification propagates to the actor and leads to ill-conditioned updates.

The first issue arises from poorly scaled critic updates. In $\nabla_\phi \mathcal{L}_{\mathrm{TD}}
= 2\mathbb{E}\left[(Q_{\mathrm{tot}}-y)\nabla_\phi Q_{\mathrm{tot}}\right]$, if $Q_{\mathrm{tot}}$ is globally rescaled by $c>1$, both the residual and Jacobian scale with $c$, and the effective Hessian scales as $c^2$. Moreover, it leads to misaligned actor gradients. The policy gradient magnitude depends on the value amplitude, \textit{i.e.}, $|\nabla_\theta \mathcal{L}_{\text{actor}}| \approx \mathbb{E}_s\left[|Q_{\mathrm{tot}}(s,\mathbf{u})||\nabla_\theta\log\pi_\theta(\mathbf{u}|s)|\right]$. Scale drift in $Q_{\mathrm{tot}}$ increases action and target variances, leading to positive feedback loop through the critic loss.

\begin{figure}[h]
    \centering
    \includegraphics[width=0.6\columnwidth]{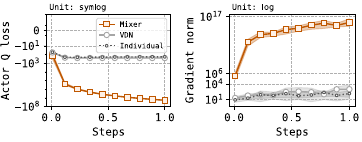}
    \caption{\textbf{Actor loss miscalibration under value-scale amplification}. (\textit{Left}) Actor loss increases sharply as value-scale drift begins, indicating that the policy objective is dominated by value amplitude rather than advantage structure. (\textit{Right}) The total gradient norm. This reveals ill-conditioned updates and confirms that the coupled actor and mixer-critic system loses numerical stability.}
    \label{fig: prob2}
\end{figure}

\textbf{Empirical results.} \textit{Figure}~\ref{fig: prob2} visualizes the empirical evidence in terms of problem II. The Q loss of actor updates decreases smoothly, but its scale rises by several orders of magnitude. Simultaneously, the total gradient norm shows exponential growth, revealing that the joint optimization becomes ill-conditioned even before explicit divergence.\footnote{The total gradient norm is computed over all parameters of actor, critic, and mixing networks.} The synchronous rise of actor loss and gradient norm demonstrates that the TD target has become a mis-calibrated supervision signal whose magnitude dominates its semantics. 

\vspace{0.1cm}
\takeawayonly{Value-scale drift miscalibrates the learning signal of the actor network, destabilizing the actor-critic update cycle through amplified gradients.}

\vspace{0.1cm}

This remark motivates the scale-invariant normalization method to stabilize the actor update when we consider value learning with non-linear value decomposition.

\begin{figure}[h]
    \centering
    \includegraphics[width=0.6\columnwidth]{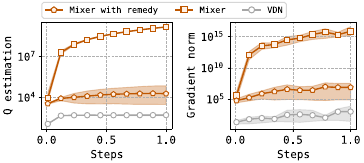}
    \caption{\textbf{Effect of the simple actor-side remedy.} These two signals together demonstrate suppression of value-scale amplification without modifying the TD objective.}
    \label{fig: remedy1}
    \vspace{-0.3cm}
\end{figure}

\textbf{Simple remedy.} We aim to prevent the actor loss from co-amplifying with the Q value, without modifying the TD objective. The idea is simple: Normalize the actor-side Q maximization term by its own batch magnitude, making the policy gradient invariant to global rescaling of $Q_{\text{tot}}$. This effectively reconditions the actor’s gradient magnitude while preserving its preference ordering. We simply modify the Q maximization of Equation~\eqref{eq: brac} as follows,
\begin{equation}
    -\frac{\mathbb{E}\big[Q_{\mathrm{tot}}(s,\mathbf u^\pi) - \mathbb{E}[Q_{\mathrm{tot}}(s,\mathbf u^\pi)]\big]}{\mathbb{E}\big[|Q_{\mathrm{tot}}(s,\mathbf u^\pi)|\big]}.
    \label{eq: actor_inv}
\end{equation}
Here, $Q_{\mathrm{tot}}(s,\mathbf u^\pi)$ denotes the critic’s estimate under the current policy. By removing the mean and dividing by the absolute mean, this operation eliminates scale drift while preserving the action preference ordering, yielding advantage-normalized and scale-invariant actor updates.

Empirically, \textit{Figure}~\ref{fig: remedy1} validates our diagnosis. Specifically, using Eq~\eqref{eq: actor_inv} suppresses the amplification of the Q value in the actor loss. The gradient norm is bounded, and the mean of $Q_{\text{tot}}$ also decreases as expected from scale normalization. These results confirm that value-scale drift is the main source of the coupled instability. Although such a remedy stabilizes the actor side, the TD objective remains scale sensitive, motivating a fully scale-invariant critic formulation.

\subsection{Scale-invariant value normalization (SVN)}
\label{subsec: fullremedy}
We now extend the invariance principle to the critic itself, ensuring that actor-critic updates become scale-invariant while preserving the theoretical foundations of TD learning.

\textbf{Desiderata.} The Bellman equation defines a fixed point under the TD operator $\mathcal TQ(s,\mathbf{u})=r+\gamma \bar{Q}(s', \pi(s'))$, that is, it minimizes $\mathbb{E}[(Q_{\text{tot}} - \mathcal TQ_\text{tot})]$. Any normalization that rescales the entire loss or prediction uniformly must therefore not change the $\arg\min$; otherwise, the Bellman fixed point would shift, violating TD.
Our goal is thus to recondition the critic updates without modifying this fixed point.

\textbf{SVN.} We compute detached statistics of the total value for each training batch: 
\begin{equation}
    \mu_Q = \text{sg}[\mathbb{E}(Q_{\text{tot}})], \quad \sigma_Q = \text{sg}[\mathrm{MAD}(Q_{\text{tot}})] + \varepsilon,
\end{equation}
where $\text{sg}[\cdot]$ denotes the stop-gradient operator and $\mathrm{MAD}(x) = \mathbb{E}[|x - \mathbb{E}(x)|]$ is mean-absolute deviation. We then define normalized projections $\widehat Q = \frac{Q_{\mathrm{tot}} - \mu_Q}{\sigma_Q} ~\text{and}~ \widehat y = \frac{y - \mu_Q}{\sigma_Q},$ and minimize the normalized TD loss as follows,
\begin{equation}
    \label{eq: svn}
    \tilde{\mathcal{L}}_{\mathrm{TD}} = \mathbb{E}\left[(\widehat Q - \widehat y)^2\right] = \frac{1}{\sigma_Q^2}\mathbb{E}\left[(Q_{\mathrm{tot}} - y)^2\right].
\end{equation}
Since $(\mu_Q, \sigma_Q)$ are treated as constants with respect to gradients, the optimization objective satisfies $\arg\min_\phi \tilde{\mathcal{L}}_{\mathrm{TD}} = \arg\min_\phi \mathcal{L}_{\mathrm{TD}},$ and thus preserves the same Bellman fixed point. In effect,  our remedy $\tilde{\mathcal{L}}_{\mathrm{TD}}$ only rescales the gradient magnitude by the batch-dependent constant $1/\sigma_Q^2$, improving the numerical conditioning of updates without altering the underlying TD solution.

To analyze its impact, we linearize the mixing function $f_{\text{mix}}(Q^1,\ldots,Q^A)$ with Jacobian $J_s$.  The effective critic Jacobian becomes $J_s^{\text{eff}} = \tfrac{1}{\sigma_Q} J_s$, which directly reduces the closed-loop gain between actor and critic:
\begin{equation}
    \tilde{g}_{\text{loop}} = \tfrac{\gamma}{\sigma_Q}\|J_s\|_{\mathrm{op}} \Big\|\tfrac{\partial \pi}{\partial Q_{\mathrm{tot}}}\Big\|.
\end{equation}
Hence, normalization can attenuate the amplification identified in Section~\ref{subsec: prob1}. By dividing out the global value scale, our remedy restores the contractive behavior of the TD operator while keeping the Bellman fixed point intact.

\takeawayonly{\texttt{SVN} retains theoretical correctness and stabilizes actor-critic-mixer coupling without modifying the Bellman fixed point.}

\begin{figure}[h]
    \centering
    \includegraphics[width=0.6\columnwidth]{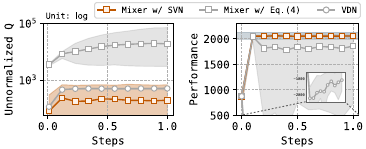}
    \caption{\textbf{SVN's effectiveness}. (\textit{Left}) Un-normalized Q value to fairly compare between each solution. (\textit{Right}) Average performance curve across $10$ evaluations $\times$ $8$ random seeds. {\color{darkUT_B}Horizontal shaded area} represents the reward distribution of dataset $\mathcal D$.}
    \label{fig: svn}
\end{figure}

\textbf{Empirical results.}\footnote{We report unnormalized Q to ensure a fair comparison between with and without \texttt{SVN}.}
\textit{Figure}~\ref{fig: svn} evaluates monotonic mixer with \texttt{SVN} and baselines to demonstrate its effectiveness. The unnormalized Q values show that the actor-only normalization partially mitigates cales drift but still allows slow amplification, whereas \texttt{SVN} completely stabilizes the value scale throughout training. Correspondingly, \textit{Figure}~\ref{fig: svn} (\textit{Right}) confirms that \texttt{SVN} maintains stable learning dynamics and achieves performance comparable to the reward distribution of the expert dataset. These results confirm that addressing the critic side of the value-scale coupling is essential for achieving fully stable actor-critic optimization.

\section{A practical recipe for offline MARL}
\label{sec: recipe}
We now broaden our scope from the theoretical analysis to an empirical analysis of the structural components of offline MARL as a whole. While Section~\ref{sec: divergent} established a theoretical foundation for stability, the practical imperative is to ensure these gains translate into robust optimization dynamics. We ask: how do different design choices interact to shape the final performance across value decomposition, value learning, and policy extraction?

To this end, we empirically study offline MARL through three modules to identify which design choices most importantly affect final performance.

\begin{figure*}[t]
    \centering
    \includegraphics[width=\textwidth]{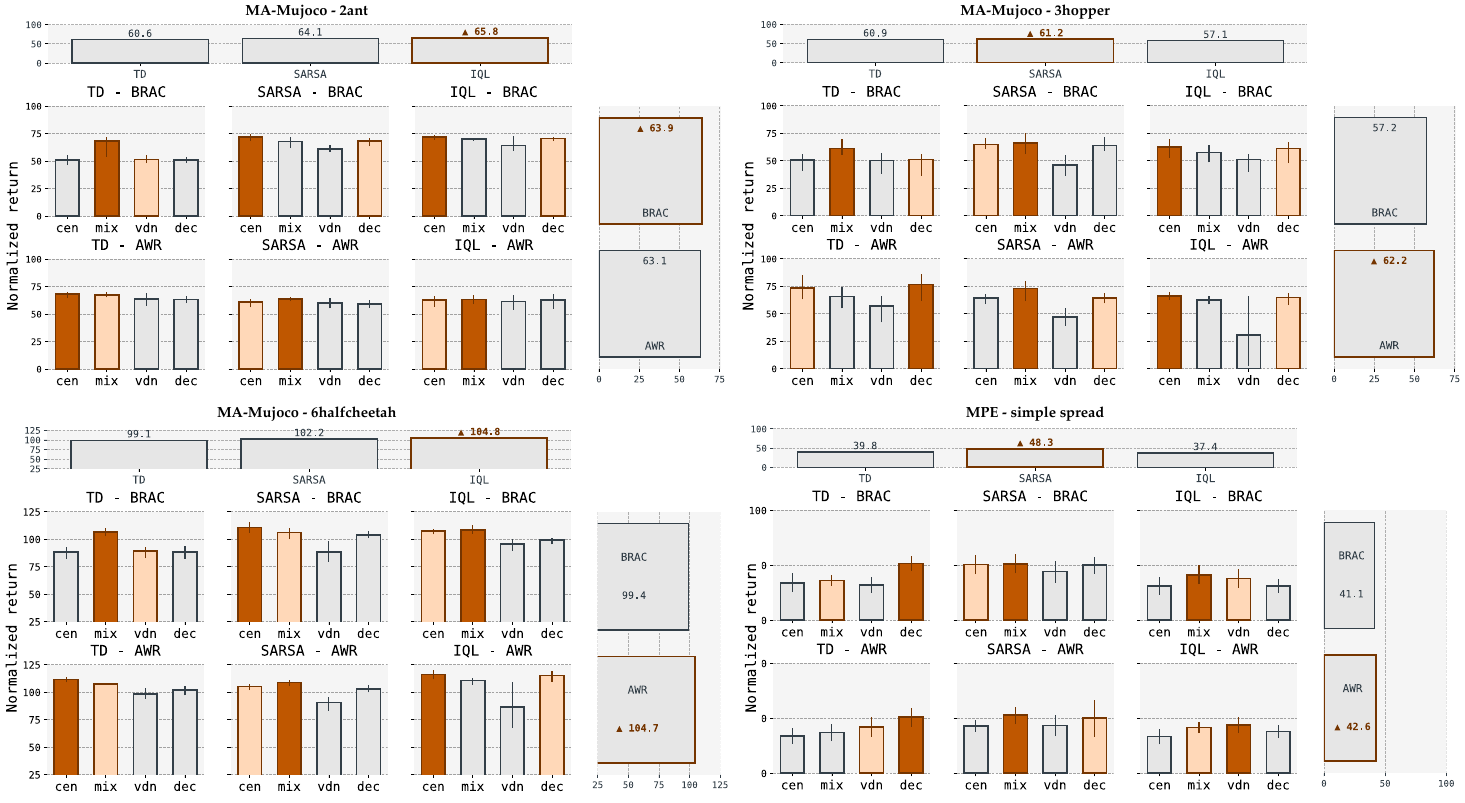}
    \caption{\textbf{Best performance according to design choices in offline MARL over four continuous control tasks.} Each bar plot reports the best normalized return over 8 seeds for different combinations of value decomposition, value learning, and policy extraction methods, aggregated across datasets and $\alpha$ hyperparameter for policy extraction. The bars colored by {\color{darkUT}dark orange} and {\color{apricot}apricot} indicate the best and runner-up performance. The top and right panels marginalize over policy extraction and value learning.}
    \label{fig: main}
\end{figure*}

\subsection{Analysis setup}
This subsection introduces the objectives for each module and the environments and datasets we study in our analysis.

\textbf{Value decomposition.} We consider four value decomposition strategies: Fully centralization (\texttt{Cen})~\citep{lyu2023centralized}, non-linear value decomposition (\texttt{Mix})~\citep{rashid2020monotonic}, linear value decomposition (\texttt{VDN})~\citep{sunehag2017value}, and fully decentralization (\texttt{Dec})~\citep{wang2022individual}.

\textbf{(\textit{1}) \texttt{Cen}}: Fully centralized critic represents the joint Q-function as a single critic network conditioned on the full global state and all agents' actions $Q_{\text{tot}}(s,\mathbf{u}) = Q(s,\mathbf{u})$. This captures all inter-agent dependencies directly.

\textbf{(\textit{2}) \texttt{Mix}}: It simply follows Equation~\eqref{eq: mixer} with Equation~\eqref{eq: svn}.

\textbf{(\textit{3}) \texttt{VDN}}: It assumes an additive structure, \textit{i.e.}, $Q_{\text{tot}}(s,\mathbf{u}) = \sum_{a=1}^A Q_a(\tau^a,u^a)$.
This is a simple linear decomposition, removing inter-agent coupling in gradients.

\textbf{(\textit{4}) \texttt{Dec}}: Fully decentralized variant does not take into account global value function $Q_{\text{tot}}(s,\mathbf{u})$, building a set of individual actors and critics. 

\textbf{Value learning.} We consider three objectives that are widely used in offline RL settings: \texttt{TD}, \texttt{SARSA}~\citep{sutton1998reinforcement}, and implicit Q learning (\texttt{IQL})~\citep{kostrikov2021offline}.\footnote{We do not consider explicit value regularization for pessimism, as such mechanisms do not naturally extend to online learning.}

\textbf{(\textit{1}) \texttt{TD}}: This follows from Equation~\eqref{eq: tdloss}.

\textbf{(\textit{2}) \texttt{SARSA}}: It is similar to the \texttt{TD} method, but we remove the policy sampling for the target calculation.
\begin{equation}
\min_{Q_{\text{tot}}}\mathbb{E}_{(s, \mathbf{u}, r, s’, \mathbf{u}') \sim \mathcal{D}} \left[ \left(Q_{\text{tot}}(s, \mathbf u) - r - \gamma \bar{Q}_{\text{tot}}(s’, \mathbf {u’})^2  \right)\right]   
\label{eq: sarsa}
\end{equation}

\textbf{(\textit{3}) \texttt{IQL}}: Unlike \texttt{TD} estimation, this implicitly emphasizes high-return actions within the behavior dataset via asymmetric regression. Therefore, it enables a behavior-constrained approximation to the Bellman optimal $\operatorname{argmax}$.
\begin{align}
    &\min_{Q_{\text{tot}}}\mathbb{E}_{(s, \mathbf{u}, r, s’) \sim \mathcal{D}} \left[\left(Q_{\text{tot}}(s, \mathbf u) - r - \gamma {V}_{\text{tot}}(s’)^2  \right)\right]  \\
    &\min_{V_{\text{tot}}}\mathbb{E}_{(s, \mathbf{u}, r) \sim \mathcal{D}}[\ell^2_\tau(Q_{\text{tot}}(s,\mathbf{u}) - V_{\text{tot}}(s)]
\end{align}
$\ell^2_\tau$ denotes the expectile regression loss, defined as an asymmetric squared loss, with expectile hyperparameter $\tau$. 

\textbf{Policy extraction.} We consider two mainstream policy extraction methods in offline RL: \texttt{BRAC}~\citep{fujimoto2021minimalist} and \texttt{AWR}~\citep{peng2019advantage}.\footnote{When extracting a policy independently of the value decomposition and learning method, we adopt a simple remedy described in Equation~\eqref{eq: actor_inv}. Other methods appear relatively stable in terms of value learning, but they sometime exhibit significant instability during the policy extraction process.} 

\textbf{(\textit{1}) \texttt{BRAC}}: This couples Q-value maximization with behavioral regularization, thereby preventing the learned policy from deviating far from the action distribution supported by the behavior policy or offline dataset. It follows the Equation~\eqref{eq: brac}, and we set $h(\cdot,\cdot)$ as the simplest regularization, which is BC loss minimization $\alpha \log \pi(u^a|o^a)$.

\textbf{(\textit{2}) \texttt{AWR}}: optimizes a weighted maximum-likelihood objective, increasing the likelihood of actions proportional to their estimated advantages $Q(o,u) - V(o)$ as follows.
\begin{equation}
    \max_{\pi_a} \mathbb{E}_{(o^a,u^a)\sim \mathcal D}[e^{\alpha(Q_a(o^a,u^a)-V_a(o^a))}\log\pi_a(u^a|o^a)]
\end{equation}

\textbf{Environments and datasets.} This work focuses on the actor-critic algorithm for multi-agent systems. Therefore, we evaluate on continuous action domains, \textit{e.g.}, MA-MuJoCo~\citep{peng2021facmac} and MPE~\citep{lowe2017multi}, as the main set of experiments. We use datasets collected from offline MARL benchmarks~\citep{formanek2023off, formanek2024dispelling}.

\subsection{Best practices for offline MARL}
\textit{Figure}~\ref{fig: main} provides a comparison across value decomposition, value learning, and policy extraction methods. It reports the best normalized return achieved for each configuration over four datasets, aggregated from 16,384 independent runs (8 seeds; Appendix~\ref{app: num}). Overall, the results reveal that performance differences are dominated by value decomposition and policy extraction, while the impact of value learning is comparatively minor.
 
For \textbf{value decomposition}, \texttt{Mix} consistently dominates the design space, achieving the best or runner-up performance in $17$ out of $24$ configurations. While fully centralized critics (\texttt{Cen}) can be competitive in certain configurations, their performance could be less consistent across design choices. Next, \texttt{VDN} exhibits a clear performance ceiling due to its restrictive additive structure. In contrast, \texttt{Mix} enables expressive modeling of inter-agent interactions. It preserves decentralized action selection and yields more consistent performance across design choices.

\begin{figure}[h]
    \centering
    \includegraphics[width=0.6\columnwidth]{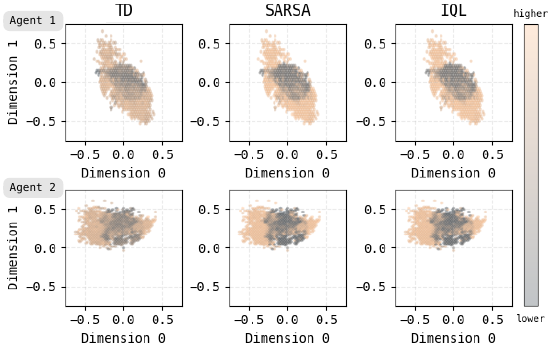}
    \caption{\textbf{Learned Q under different value learning methods.} Each point represents a dataset action sample, colored by its Q value: {\color{darkUT_B}dark blue} and {\color{apricot}apricot} indicate lower and higher Q values. The $x$ and $y$ axes are the coordinates of dimension $0$ and $1$ of action. The top and bottom are about agents $1$ and $2$.}
    \label{fig: q_}
    \vspace{-0.2cm}
\end{figure}

For \textbf{value learning} objectives, we observe that the objectives that avoid policy-sampled target estimation (\textit{i.e.}, \texttt{SARSA} and \texttt{IQL}) tend to be slightly more favorable than \texttt{TD} in the offline setting. This is because \texttt{SARSA} and \texttt{IQL} can provide more conservative target estimates. However, the performance differences among these objectives are modest and do not constitute a dominant factor compared to value decomposition and policy extraction.

This observation is supported by \textit{Figure}~\ref{fig: q_}. \texttt{TD}, \texttt{SARSA}, and \texttt{IQL} exhibit highly similar value estimation behavior on in-distribution samples. The learned Q-value distributions largely overlap in action space, with comparable separation between low- and high-value regions, indicating that all three methods capture similar relative preferences over dataset actions. Notably, \texttt{TD} exhibits slightly reduced contrast in Q-value, with fewer sharply distinguished high-value points suggesting weaker relative value separation. However, this difference does not translate into substantial performance gaps once value decomposition and policy extraction are fixed, reinforcing the conclusion that value learning itself is not the primary bottleneck in offline MARL.

For \textbf{policy extraction}, \texttt{AWR} yields more stable and reliable results than \texttt{BRAC} across value decomposition and value learning choices. In our observation, while \texttt{BRAC} occasionally attains strong performance on expert datasets, it frequently suffers from sharp degradation, likely due to its mode-seeking behavior inducing out-of-distribution joint actions. In contrast, \texttt{AWR}’s mode-covering nature better preserves coordinated behavior, particularly when paired with the non-linear value decomposition (\texttt{Mix}).

\section{Discussion and further analysis}

\begin{figure}[h]
    \vspace{-0.2cm}
    \centering
    \includegraphics[width=0.6\columnwidth]{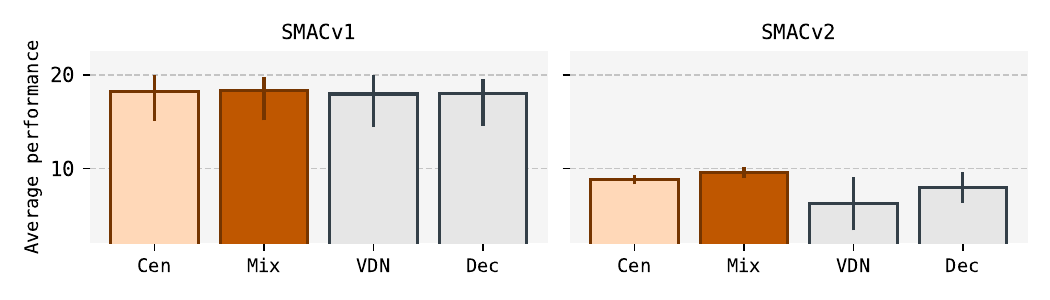}
    \caption{\textbf{Performance comparison on discrete control.} We evaluate four tasks from SMACv1 (Good and Medium dataset \texttt{3m} and \texttt{2s3z}) and two tasks from SMACv2 (Replay dataset for \texttt{terran\_5\_vs\_5} and \texttt{zerg\_5\_vs\_5}). The error bar shows the minimum and maximum performance range.}
    \label{fig: discrete}
\end{figure}

\textbf{Do These Ideas Work on Discrete Control?}
Beyond continuous control, we examine whether the efficiency of \texttt{Mix} extends to discrete control, \textit{i.e.}, SMACv1~\citep{samvelyan2019starcraft} and SMACv2~\citep{ellis2023smacv2}. As shown in \textit{Figure}~\ref{fig: discrete}, \texttt{Mix} consistently outperforms the other value decomposition methods in discrete settings. Specifically, while all methods achieve comparable performance on SMACv1, \texttt{Mix} demonstrates superior performance on SMACv2. This suggests that \texttt{Mix} is particularly effective in environments characterized by high stochasticity. 

\begin{figure}[h]
    \centering
    \includegraphics[width=0.6\columnwidth]{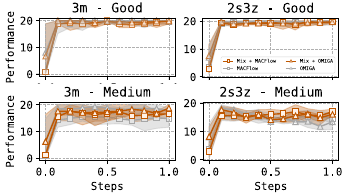}
    \caption{\textbf{Compatibility of \texttt{Mix}}. We integrate \texttt{Mix} into two offline MARL algorithms, \texttt{MAC-Flow} and \texttt{OMIGA}, and compare their performance against the original baselines on SMACv1 tasks.}
    \label{fig: prac}
\end{figure}

\textbf{Can Mixer Be Incorporated into Prior Algorithms?} 
To verify the practicality of non-linear value decomposition (\texttt{Mix}), we integrate it with recent algorithms, \texttt{MAC-Flow}~\citep{lee2025multi} and \texttt{OMIGA}~\citep{wang2023offline}, and check whether this integration can further boost performance with these methods. On a good-quality dataset, replacing the value decomposition with \texttt{Mix} maintains the existing superior performance without degradation. More importantly, on a suboptimal (meidum) dataset, integrating \texttt{Mix} can enhance the baselines' performance.

\begin{figure}[h]
    \centering
    \includegraphics[width=0.8\columnwidth]{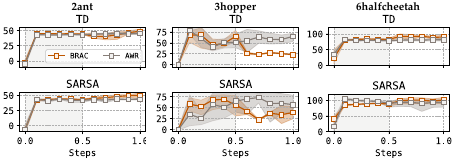}
    \caption{\textbf{Performance in offline-to-online MARL.} Online fine-tuning starts at $0.5$ gradient steps normalized by $[0,1]$ scale. }
    \label{fig: o2o}
\end{figure}

\textbf{How does \texttt{Mix} improve with additional interaction data?} 
\textit{Figure}~\ref{fig: o2o} shows that \texttt{BRAC} frequently benefits from online interaction but also occasionally suffers from performance degradation. On the other hand, \texttt{AWR} largely preserves its offline performance after online fine-tuning. This observation suggests that the effect of online fine-tuning in MARL is sensitive to the underlying policy mode rather than uniformly beneficial. Overall, we highlight the need for online fine-tuning strategies specifically designed for MARL, beyond direct adaptations of offline MARL methods.

\closing{Our empirical study demonstrates that stabilized non-linear value decomposition is effective across continuous and discrete control. Additionally, we find the importance of mode-covering policy extraction to preserve coordination patterns in offline MARL.}

\section{Call to actions: Towards practical and scalable offline MARL}
In this work, we empirically demonstrated that the key to offline MARL is how the policy is extracted to preserve coordination patterns and how the global value is made from individual value functions. Additionally, this work analyzed the problem of non-linear value decomposition and provided some simple but powerful remedies. This is far from the existing trend of offline MARL, which heavily focuses on the extension of value regularization of single-agent offline RL. Overall, this work repositions non-linear methods from a fragile component to a foundational building block for scalable and practically deployable offline MARL by providing both a diagnostic understanding and a recipe for offline MARL. 

Although this is a great start, further research is needed to build upon these findings. First, our approach relies on simple normalization methods to stabilize the scale of value from adoption to non-linear value decomposition. This implicitly assumes that such scale control is sufficient. This leaves open the need to more directly develop non-linear value decomposition itself, including alternative architectural designs and stabilization principles beyond normalization. Second, while our study adopts the same action discretization method across all combinations, the optimal one may depend on the specific combination of value decomposition, value learning, and policy extraction. Since actor-critic structures are primarily designed for continuous control, the impact of different discretization choices in offline MARL warrants more systematic investigation. Finally, our empirical setup largely depends on the classical MARL testbed, \textit{i.e.}, MA-MuJoCo, SMAC, and MPE, which considers team rewards as individual ones in dense reward setups. This posits relatively weak non-linear coupling among agents, as well as making it hard to study scaling behavior with substantially larger datasets.

Such shortcomings serve as a springboard for fruitful research trajectories in offline MARL:

\begin{itemize}
    \item How should we completely stabilize the hypernetwork for non-linear value decomposition (\textit{e.g.}, potentially with a dueling mechanism~\citep{sutton1998reinforcement, wang2016dueling, wang2020qplex}, attention structure~\citep{vaswani2017attention, yang2020qatten}, and factored graph~\citep{guestrin2001multiagent, guestrin2002coordinated, bohmer2020deep, kang2022non})?
    \item Can offline MARL benchmarks and datasets be expanded to better capture diverse coordination structures, such as goal-conditioned tasks~\citep{feng2025safe, skrynnik2024pogema, lee2025learning}, skill-based coordination~\citep{liu2022heterogeneous, omari2025multi, chen2022scalable}, and social or mixed cooperative-competitive settings~\citep{guo2025socialjax, ruhdorfer2024overcooked, gessler2025overcookedv2}, beyond current dense team-reward testbeds?
    \item Are the scalability limits of offline MARL fundamental, or can principled designs enable reliable offline-to-online MARL (or scaling up the datasets in offline)~\citep{park2024value, lee2022offline, sun2020scaling}?
\end{itemize}

\section*{Impact statement}
This work advances offline multi-agent reinforcement learning by improving the stability of value decomposition and multi-agent policy extraction, and contributes to the broader field of machine learning.

\bibliography{CITE}
\bibliographystyle{style}

\newpage
\appendix

\par\noindent\rule{\textwidth}{2pt}
\begin{table}[h]
    \vspace{-0.3cm}
    \centering
    \Large
    \begin{tabular}{c}
\textbf{Appendix}
    \end{tabular}
    \vspace{-.5cm}
\end{table}
\par\noindent\rule{\textwidth}{1pt}

\tableofcontents
\addtocontents{toc}{\protect\setcounter{tocdepth}{2}}

\newpage

\section{Miscellaneous}
\subsection{Summary of notations}
\begin{table}[h]
    \centering
    \resizebox{\textwidth}{!}{
    \begin{tabular}{c l c l}
        \multicolumn{4}{c}{\texttt{Dec-POMDP elements}} \\
        \toprule
        \textbf{Notation} & \textbf{Description} & \textbf{Notation} & \textbf{Description} \\
        \midrule
        $\mathcal{A}$ & set of agents & $a$ & agent index \\
        $A$ & number of agents & $\gamma \in [0,1)$ & discount factor \\
        $\mathcal{S}$ & global state space & $s$ & global state \\
        $\mathcal{O}_a$ & observation space of agent $a$ & $o^a$ & local observation of agent $a$ \\
        $\mathcal{U}_a$ & action space of agent $a$ & $u^a$ & action of agent $a$ \\
        $\mathcal{P}$ & state transition function & $\Omega$ & observation function \\
        $r_a$ & reward function of agent $a$ & $R$ & team reward \\
        $\tau^a$ & trajectory of agent $a$ & $\mathcal{D}$ & offline dataset (replay buffer) \\
        \bottomrule
        \\
        \multicolumn{4}{c}{\texttt{Algorithm elements}} \\
        \toprule
        \textbf{Notation} & \textbf{Description} & \textbf{Notation} & \textbf{Description} \\
        \midrule
        $Q_\text{tot}$ & global state-action value function & $Q^a$ & state-action value function of agent $a$ \\
        $f_\text{mix}$ & mixing function & $h(\cdot, \cdot)$ & regularization function \\
        $\pi$ & offline policy & $\mu$ & behavioral policy \\
        \bottomrule
        \\
        \multicolumn{4}{c}{\texttt{RL Training}} \\
        \toprule
        \textbf{Notation} & \textbf{Description} & \textbf{Notation} & \textbf{Description} \\
        \midrule
        $\theta$ & policy parameters & $\bar{\theta}$ & target critic parameters \\
        $\phi$ & critic parameters & $\phi_\text{mix}$ & mixing network parameter \\
        \bottomrule
    \end{tabular}}
\end{table}

\section{Extended related works}
\subsection{Actor critic with mixing network}
Mixing networks were initially formulated for value-based discrete control, where Q-functions implicitly define policies. Recent research has adapted these architectures to actor-critic frameworks to focus on structured credit assignment~\citep{su2021value, peng2021facmac, hu2021rethinking, ding2025multi, ji2025cora}. This integration proves essential for centralized training when decoupling policy optimization from value estimation. By aggregating agent-wise utilities into a global objective, mixing-based critics preserve the coordination benefits of factorization. This paradigm effectively combines scalable value decomposition with the flexibility of independent policy learning.

Compared to fully centralized critics that directly condition on the joint observation and joint action space, actor-critic methods with mixing networks offer improved scalability as the number of agents increases. Fully centralized value functions suffer from exponential growth in input dimensionality and often become impractical in environments with many agents or high-dimensional observations~\citep{lirevisiting, lyu2023centralized, lyu2021contrasting, foerster2017stabilising}. In contrast, mixing-based critics decompose the value estimation into per-agent components that are aggregated through a structured mixing function, enabling parameter sharing and modularity while retaining access to global state information during training.

However, existing methods with non-linear value decomposition have largely been studied in an online setting. Much less is understood about how non-linear value decomposition behaves in offline MARL. In particular, it remains unclear how the current RL recipe with value decomposition methods~\citep{park2024value, wu2019behavior, peng2019advantage, wang2020qplex, son2019qtran, rashid2020monotonic, rashid2020weighted} interacts with OOD issues and coordination errors under offline training. Therefore, in this work, we investigate how non-linear value decomposition can be effectively incorporated into offline MARL.

\subsection{Non-linear value decomposition}
The core challenge in MARL is credit assignment, which has motivated value decomposition methods that structure global value estimation into agent-wise components. Early work, \textit{e.g.}, VDN~\citep{rashid2020monotonic}, assumes a linear additive structure over agent-wise value functions. This enables scalable learning but limits the expressivity of the global value function. QMIX~\citep{rashid2020monotonic, rashid2020weighted} extends this formulation by introducing a state-conditioned mixing network, allowing non-linear yet monotonic aggregation of individual Q-values and improving its expressivity. QTRAN~\citep{son2019qtran} and other variants~\citep{wang2020qplex, yang2020qatten, bohmer2020deep, peng2021facmac, wang2020roma}, \textit{e.g.}, graph-based critics and attention mechanisms, further relax structural constraints to model more general non-linear interactions among agents.

Despite this progress, existing non-linear value decomposition methods have been exclusively studied in online settings. In offline MARL, the behavior of non-linear mixing-based critics remains largely unexplored; to the best of our knowledge, there has been no systematic investigation of non-linear value decomposition in an offline setting. This gap motivates our study, which examines how mixing network-based critics can be incorporated and analyzed in offline multi-agent settings, and what design choices are necessary to retain their scalability and coordination benefits under offline constraints.

\section{Experimental Details}
\subsection{Two-step coordination game}
We consider a simple two-step cooperative matrix game designed to highlight the limitations of linear value decomposition and the necessity of a non-linear mixing network. 

\textbf{Game description}. The MDP of this game consists of two agents, three states, and binary actions, and unfolds over two timesteps. At the first timestep, the environment is in an initial state $s_1$, where Agent A selects the next state (Agent B chooses an action, but it does not affect the state change). In particular, Agent A chooses between two options: (\textit{i}) a safe state $s_{2-1}$, which leads to a deterministic but suboptimal outcome, and (\textit{ii}) a risky state $s_{2-2}$, which enables a higher optimal reward but requires coordinated behaviors. The reward structures of each state under coordinated joint actions are given as follows:

$$
R(s_{2\text{-}1}) =
\begin{pmatrix}
7 & 7 \\
7 & 7
\end{pmatrix},
\qquad
R(s_{2\text{-}2}) =
\begin{pmatrix}
0 & 1 \\
1 & 8
\end{pmatrix}.
$$

\textbf{Why linear value decomposition fails?} \texttt{VDN} assumes that the joint action-value function is a linear sum of individual agent values. This assumption is violated in the risky state $s_{2-2}$, where the optimal reward arises only from a specific coordinated joint action. Because the benefit of coordination cannot be attributed additively to individual agents, VDN underestimates the value of the risky state. As a result, the agent selecting the state transition prefers the safe but suboptimal state $s_{2_1}$, leading to a suboptimal joint policy.

\subsection{Normalized score for continuous control benchmarks}
To enable consistent comparison across different continuous control benchmarks, we report normalized scores using $\min-\max$ normalization. For each task, the normalized score is computed as
\[
\text{Normalized Score} =
\frac{J(\Pi) - \text{scale}_{\min}}
{\text{scale}_{\max} - \text{scale}_{\min}},
\]
where $J(\Pi)$ denotes the average episode return of the evaluated set of policies, and $\text{scale}_{\min}$ and $\text{scale}_{\max}$ define the task-specific minimum and maximum reference returns.

For the MA-MuJoCo benchmarks, $\text{scale}_{\max}$ is defined as the maximum return achieved by the Expert trajectories (the best quality dataset) in the dataset, while $\text{scale}_{\min}$ corresponds to the minimum return observed in the Medium-Replay dataset (the lowest quality dataset). Specifically, we use the following values:
\begin{align}
\text{2ant:} \quad & \text{scale}_{\min}=895.37,\quad \text{scale}_{\max}=2124.15, \nonumber \\
\text{3hopper:} \quad & \text{scale}_{\min}=70.75,\quad \text{scale}_{\max}=3762.68, \nonumber \\
\text{6halfcheetah:} \quad & \text{scale}_{\min}=-198.76,\quad \text{scale}_{\max}=3866.08. \nonumber
\end{align}
For the MPE benchmark, we follow previously reported evaluation protocols and adopt the reference values reported in prior work~\citep{lee2025multi, formanek2023off, formanek2024dispelling}. In particular, for the Spread task, we use the following scale values:
$$\text{scale}_{\min}=159.8,\quad \text{scale}_{\max}=516.8.$$. 

\subsection{The number of experiments for \textit{Figure}~\ref{fig: main}}
\label{app: num}
For \textit{Figure}~\ref{fig: main}, we run $16,384$ independent runs with hyperparameter sweeps. Specifically, this corresponds to sweeping over the Cartesian product of hyperparameters for each learning algorithm: for \texttt{TD} and \texttt{SARSA}, we consider $4$ value decomposition methods, $2$ policy extraction methods, $4$ policy extraction temperatures $\alpha$, $4$ tasks, $4$ datasets per task, and $8$ random seeds, resulting in $4,096$ runs each; for \texttt{IQL}, we additionally sweep over $2$ expectile loss coefficients $\tau$, yielding $8,192$ runs. In total, this amounts to $16,384$ runs.

\section{Implementation details}
\textbf{Git repository.} We provide our codebase implementation on \url{https://github.com/DongsuLeeTech/offline-marl-recipe}

\textbf{Actor-critic network architecture.} Both the actor and critic are parameterized by multi-layer perceptrons (MLPs) with four hidden layers of size $[512,512,512,512]$. The actor outputs continuous values, while the critic is implemented as a double Q-network with two ensemble heads~\citep{van2016deep}. Layer normalization~\citep{ba2016layer} is applied to all critic layers. Target networks are maintained for both the critic and the mixing network, and are updated using Polyak averaging~\citep{polyak1992acceleration} with coefficient $\tau$. 

\textbf{Action discretization.} For discrete control domains, actions are represented internally using one-hot vectors from the continuous value of the actor network~\citep{wu2019behavior}. The actor predicts logits over the discrete action space. During evaluation, actions are selected via argmax over the actor outputs.

\textbf{Mixing network.} This is a hypernetwork-based architecture that generates state-dependent mixing weights~\citep{rashid2020monotonic}. In our implementation, the agent-wise Q-values are first embedded into a $32$-dimensional latent space, and the hypernetwork uses a hidden dimension of $128$ to produce the mixing coefficients from the global state. The mixing network operates on per-agent Q estimates from all individual critics and is trained jointly with the critic. A separate target mixing network is maintained and used for computing TD targets to improve training stability~\citep{qin2024dormant}.

\textbf{Scale-invariant value normalization.} We apply a scale-invariant value normalization scheme when training the critic with a mixing network. Specifically, the TD loss is computed using normalized Q-values, where both current and target Q-values are centered by the mean and scaled by the mean absolute deviation of the current Q estimates. The normalization statistics are detached from the gradient graph to preserve Bellman invariance. This normalization is applied only to the critic loss and does not affect policy evaluation or execution. We provide Python-style pseudocode of \texttt{SVN} below.

\begin{figure}[h]
\UsePyMaterialLight
\begin{lstlisting}[caption={Python style pseudocode of scale-invariant value normalization.}, label={lst:svn}]
### SVN (Scale-invariant Value Normalization) for TD learning

# Inputs:
#   qs[0], qs[1]   : ensemble total Q estimates
#   targets        : Bellman targets

q_tot_curr = minimum(qs[0], qs[1])
target_q   = stop_gradient(targets)

# Detached normalization statistics computed only from the current total Q
mu_q  = stop_gradient(mean(q_tot_curr))
mad_q = stop_gradient(mean(abs(q_tot_curr - mu_q))) + eps

# Normalize both current Q and target
q_hat = (q_tot_curr - mu_q) / mad_q
t_hat = (target_q   - mu_q) / mad_q

critic_loss = mean((q_hat - t_hat)**2)
\end{lstlisting}
\end{figure}

\textbf{Online fine-tuning.} For the offline-to-online experiments, we deviate from the common practice of \textit{balanced sampling}, which incorporates offline data during online training~\citep{ross2012agnostic, ball2023efficient, eom2024selective, lee2025scenario}. Instead, our approach focuses exclusively on newly collected online rollouts. Starting from the offline pretraining checkpoint, the agent undergoes an additional $500$K gradient steps of pure online training.

\textbf{Training and evaluation.} We train all methods with $1$M gradient steps for SMACv1 and SMACv2, and $500$K steps for MPE and MA-MuJoCo. For offline-to-online training, we first perform $500$K steps of offline training, followed by $500$K steps of online training. We evaluate the learned policy every $50$K steps using $10$ evaluation episodes. 




\end{document}